Authors: Bettina Berendt and Stefan Schiffner

# Whistleblower protection in the digital age - why "anonymous" is not enough. From technology to a wider view of governance.

**Abstract:**

When technology enters applications and processes with a long tradition of controversial societal debate, multi-faceted new ethical and legal questions arise. This paper focusses on the process of whistleblowing, an activity with large impacts on democracy and business. Computer science can, for the first time in history, provide for truly anonymous communication. We investigate this in relation to the values and rights of accountability, fairness and data protection, focusing on opportunities and limitations of the anonymity that can be provided computationally; possible consequences of outsourcing whistleblowing support; and challenges for the interpretation and use of some relevant laws. We conclude that to address these questions, whistleblowing and anonymous whistleblowing must rest on three pillars, forming a "triangle of whistleblowing protection and incentivisation" that combines anonymity in a formal and technical sense; whistleblower protection through laws; and other norms and practices including organisational error culture.

**Authors:**

Prof. Dr. Bettina Berendt:

- TU Berlin and Weizenbaum Institute, Germany, and KU Leuven, Belgium;
  Hardenbergstr. 32, D-10623 Berlin, Germany
- ✉ bettina.berendt@kuleuven.be, 💻 www.berendt.de/bettina

Dr. Stefan Schiffner

- University of Münster, Germany;
  Einsteinstraße 64, D-48149 Münster, Germany
  ✉ Stefan.Schiffner@uni-muenster.de, 💻 https://www.uni-muenster.de/Informatik/u/schiffner/

# 1 Introduction

In an age of smart systems, some of the most interesting ethical challenges arise when technology enters *application areas* and *processes* that have a long tradition already fraught with multi-faceted ethical questions, involving infringements of *values* and/or *rights*. Even where the use of computing *methods* can improve the functioning of these systems, risks abound: engineers may perceive the chance to "solve" a complex social problem; businesses may oversell the capabilities of technology; legal actors may find themselves confronted with legal regulations based on an incomplete understanding of technology, or trying to craft new regulation that is sufficiently technology-agnostic to be effective in rapidly-changing environments; and social scientists may feel stunted in their efforts to point to the "solutions" risking to perpetuate or even worsening existing injustices. Bridges must be built between these understandings and also to the pertinent insights by other stakeholders.

In this paper, we focus on the *process* of whistleblowing. Whistleblowing occurs in many *application areas*, it is an activity of those with less power, and one that is increasingly being recognised as essential for democracy and even businesses to function well. In contrast to other areas of computerisation (such as large-scale data gathering and processing, or automated decision-making), the spectre here is not automation. Whistleblowing often requires tremendous amounts of perceptiveness, critical reasoning, moral deliberation and courage, in short, human intelligence. Therefore, IT-based *methods* are regarded as a game-changer predominantly with regard to how they can support communication – specifically, in how computer science can, for the first time in history, provide for truly anonymous communication. This matters because the possibility to report anonymously is considered a major incentive for whistleblower to come forward. We will investigate how anonymous reporting affects the *values and rights* of accountability, fairness and data protection (in addition to the many different values upheld by people reporting wrongdoing: health, safety, worker's rights, non-discrimination, non-corruption, and many others). The effects of whistleblowing on these values and rights are often most acutely felt by the whistleblowers, who act out of a position with relatively little power compared to those they report on (Westin, 1981, pp.1, 2), but in general distributed over many stakeholders.

Some of these rights have a legal basis, but all of them involve ethical choices and reasoning. We briefly review, in Section 2, well-known ethical risks and dilemmas of whistleblowing in general. We then derive three new ones that arise with modern approaches to providing for computer-supported anonymous whistleblowing:

- the opportunities and limitations of the anonymity that can be provided computationally. Specifically, we start with a computer-networks perspective and (1) describe the communications problem of reporting and how state-of-the-art computational architectures and algorithms can create anonymous channels (Section 3.1), (2) show that non-technical content properties of the reporting message (which is sent over the channel) can still lead to de-anonymisation and what computational choices could do to mitigate (but not solve) this challenge of epistemic non-anonymisability (Section 3.2).
- the possible consequences of outsourcing whistleblowing support (Section 3.3),
- the challenges this poses for the interpretation and use of some relevant laws (Section 3.4),
- the question how to supplant anonymisation (which cannot be formally guaranteed) and legal protection (which does not always deter retaliation) by organisational culture (Section 3.5).

We conclude in Section 4 that in order to address these ethical challenges, whistleblowing and anonymous whistleblowing must rest on three pillars, which we call the "triangle of whistleblowing protection and incentivisation" that combines approaches and methods from at least three fields: (a) anonymity in a formal and technical sense, (b) whistleblower protection through (whistleblowing and other) laws and (c) other norms and practices including codes of conduct and organisational error culture. Such collaboration can protect different stakeholders, further economic goals, and serve the Common Good.

## 2 Whistleblowing: some key properties and recent findings

"Whistle blowers […] are employees who believe their organization is engaged in illegal, dangerous, or unethical conduct. Usually, they try to have such conduct corrected through inside complaint, but if it is not, the employee turns to government authorities or the media and makes the charge public." (Westin, 1981, p.1). Whistleblowing *laws* protect only certain types of reporting; for example, the EU Directive 2019/1937 defines whistleblowers as "reporting persons working in the private or public sector who acquired information on breaches in a work-related context" (Article 4(1)), and the breaches as violations of EU Union law as listed in its Article 2(1)(a) and Annex. Since specific laws often have a more general influence on subsequent jurisdiction and law-making, and since we are concerned with the more general (and, in our opinion, generalizable) ethical questions, we will ignore these details here.

The phenomenon is receiving growing attention: "If the twentieth century was the age of civil disobedience, the twenty-first century is shaping up to be the age of whistleblowing" (Delmas, 2015). But while there is growing consensus that company whistleblowers not only help companies detect misconduct but also save them enormous amounts of money (e.g., Hauser, Hergovits, & Blumer, 2019), and that government whistleblowers can help keep up legality and ethics of the public sector and save the taxpayer money, both private-sector and public-sector whistleblowers experience substantial or even growing pushback and retaliation (e.g. Gawlik v. Liechtenstein, see Gesley, 2021; Radack & McClellan, 2011).

From the outset, two fundamental dilemmas, equipped with further sub-dilemmas, emerge. First, the potential whistleblower has to decide between "loyalty and dissent" towards their organisation (Westin, 1981), where the need to dissent generally arises from loyalty to higher values, the common good, and often the law. A large body of literature has examined factors that favour one or the other of these choices (Butler, Serra, & Spagnolo, 2020; Park & Lewis, 2019; Lipman, 2012). Second, organisations and lawmakers, while intent on encouraging whistleblowing that in the end profits themselves and/or society, want to discourage abuses in the form of denunciation. While studies of whistleblowing suggest that the proportion of abusive reports is small (Hauser et al., 2019), the power to define what is abusive may reside with the organization, which can also be tempted to claim damages to reputation, national security, etc. and argue that these outweigh the good-faith concerns of the whistleblower or even the undisputedly unethical or illegal activities that were reported (e.g. morphine killings, dragnet surveillance, or war crimes, cf. Article 19, 2021; Satter, 2020; Coliver, 2013). Viewed from this angle, even retaliation can be re-defined as the justified holding-to-account of the whistleblower.

Regardless of whether one regards retaliative measures as justified or not, the fear of retaliation is a major disincentive for potential whistleblowers. Therefore, to the extent that whistleblowing is considered socially beneficial (see the scope of the EU Directive above as an example), laws are being created to reduce the likelihood of this grave threat. Two approaches to protect whistleblowers are currently discussed prominently: to make retaliation measures such as job termination illegal, and to enable whistleblowers to report anonymously (with the aim of making it impossible to target them). The illegality of retaliation is a key concept in all "whistleblower protection laws" such as the US Sarbanes-Oxley Act of 2002 or the European Union "Whistleblowing Directive" (Directive (EU) 2019/1937 of the European Parliament and of the Council of 23 October 2019 on the protection of persons who report breaches of Union law (European Union, 2019), henceforth WBD), directly applicable as of 2021 but effective only via each Member State's transposition into national law (which as of this writing has not happened in any country, is "in progress" in 22 Member States and has not started in 5[1]).

---

[1] https://www.whistleblowingmonitor.eu/ (retrieved 2021-08-23)

Anonymous reporting has a different legal status in different laws. The US Sarbanes-Oxley Act requires companies to provide an anonymous reporting channel. The EU WBD allows but does not require this[2] – and it is not yet clear how the different Member States will treat this aspect in their respective transpositions into national law. The question whether anonymous reporting is "ethical" has been debated for centuries if not milennia, and it remains controversial (see Elliston, 1982, for an ethics-centric analysis that results in an affirmative answer).

Reported misdeeds can not only be illegal or unethical with regard to the common good, they can also constitute grave unfairness to the employee (who is in general much less powerful than the organization). Striking examples of discrimination and harassment at work were reported by Westin (1981) and, forty years later, in the many cases around the #metoo debate (cf. Wikipedia Contributors, 2021) and further areas (Bazzichelli, 2022). The employee seeking to right the wrong and to hold the more powerful to account for it, seeks to redress this unfairness. The reported-on or the organization may retaliate "because they can", but also because *they* feel wronged and harmed. Anonymous reporting, viewed from this angle, unfairly deprives them of the possibility to hold the reporting employee to account. But does a reported-on individual have an ethical right to confront the reporter? Elliston (1982, pp.52f.) argues that this is not the case: "individuals have a right to protect themselves against false accusations that can ruin their careers and compromise their good name. But to guarantee this right, the identity of whistleblowers need not be known; it is only necessary that accusations be properly investigated, proven true or false, and the results widely disseminated. […] The taunt of the accused that the whistleblower come forward and 'face him like a man' is a bully's challenge when issued by the powerful."

## 3 Whistleblowing in the Digital Age

To describe the opportunities and limits of anonymisation, we model the basis setting of whistleblowing in terms of communications. One person A watches another person B do something that they (A) consider wrong (henceforth "the wrong"). Alternatively, A may perceive institutional arrangements or some other structure that leads to wrongs. The actor-at-fault could be human or not (e.g., an algorithmic system with effectors). A then has, roughly, three options: do nothing, intervene directly in some way to undo or modify the wrong, or report it to others in the hope that these intervene. Variations and combinations are possible. For example, it may be impossible to undo the wrong (e.g., a large-scale environmental pollution, death), and the intervention may serve other purposes (e.g., prevention of repetition, but also revenge or ill will). Also, A may try to intervene directly and report, or report to different others. A simple version of the basic set-up of reporting is shown in Fig. 1 (a). In general, the goal is to hold the wrongdoer to account (see Fig. 1 (b)), see Raab (2012) for this notation of accountability.

In the remainder of this section, we concentrate on strategies for mitigating the risk of retaliation as a central element of creating incentives for whistleblowing. We distinguish between three strategies for discouraging retaliatory behaviour: making it impossible (via technology, Sections 3.1-3.3), making it forbidden (especially

---

[2] "Without prejudice to existing obligations to provide for anonymous reporting by virtue of Union law, this Directive does not affect the power of Member States to decide whether legal entities in the private or public sector and competent authorities are required to accept and follow up on anonymous reports of breaches." (Article 6(2)).

"Without prejudice to existing obligations to provide for anonymous reporting by virtue of Union law, it should be possible for Member States to decide whether legal entities in the private and public sector and competent authorities are required to accept and follow up on anonymous reports of breaches which fall within the scope of this Directive. However, persons who anonymously reported or who made anonymous public disclosures falling within the scope of this Directive and meet its conditions should enjoy protection under this Directive if they are subsequently identified and suffer retaliation." (Recital 34)

Given that anonymous whistleblowing remains controversial in some Member states, this openness may cause misalignment across the EU and thereby weaken public policy.

via laws, Section 3.4, but also via ethics codes, Section 3.5), and making it unattractive (via organizational choices, Section 3.5). Our focus is on the technological opportunities and limitations, but our conclusion is that a combination of all governance strategies is needed for effectively incentivizing and protecting whistleblowers.

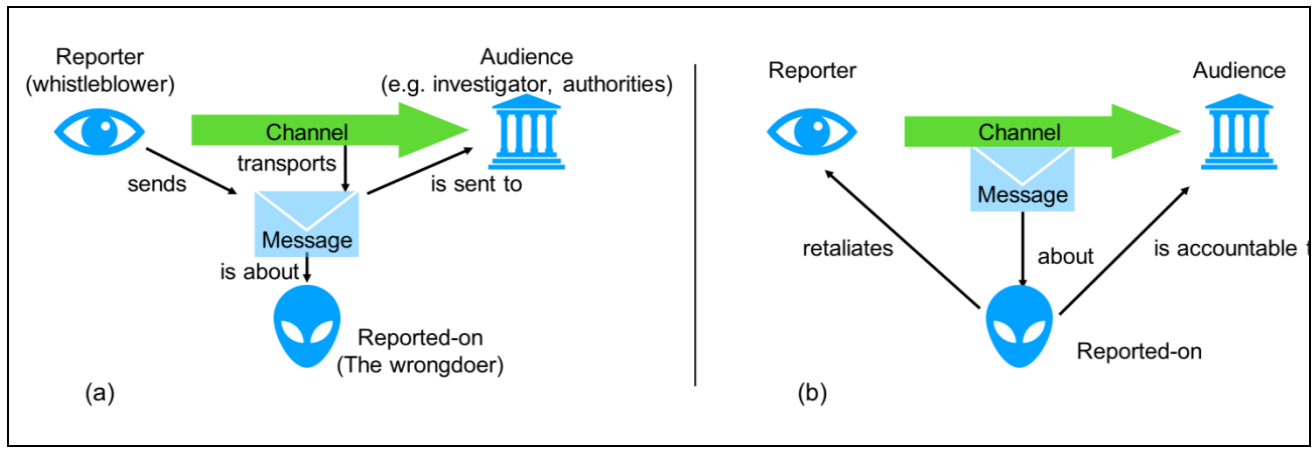

Figure 1: (a) Roles and actions in the basic reporting setting. (b) Accountability of the wrongdoer as a goal, retaliation as a possible consequence.

### 3.1 Anonymous communications for whistleblowing support

The terms anonymity, pseudonymity (the use of a pseudonym) and confidentiality are used ambiguously in everyday language. We will build on the more rigorous and widely used definitions by Pfitzmann and Hansen (2010), and we will focus on anonymous communications (rather than data in general).

If a natural person called Jane Doe (we assume this is a unique name) wants to blow the whistle via a message M, and this produces the knowledge (in someone's head) and/or data that "Jane Doe said M", she is clearly not anonymous. If all that becomes known is that "someone said M", it appears that Jane is anonymous.

However, of which "someone"s do we speak here? All human beings who ever lived? All those living now? Only all workers of Jane's company or other persons who are in contact with it in the context of their work-related activities (who could reasonably be expected to be the possible whistleblowers)? In fact, here and in general "*anonymity* of a subject means that the subject is not identifiable within a set of subjects, the anonymity set." (Pfitzmann & Hansen, 2010, p.9).

The anonymity consists in the impossibility to *link* Jane Doe (computationally, her name or other *identifier*) to M. Is it then sufficient to strip data of identifiers to make them anonymous? A typical operation is to replace person names or IP addresses by different strings or numbers (e.g. by computing a hash function value, or by using a fake name proposed by Jane). However, in most cases these new values constitute an (indirect) identifier: combined with the right background knowledge (whether in a look-up table stored in some database, computable via the hash function, etc.), the link between Jane Doe and M can be re-established. Thus, even if it is a fake name, a *pseudonym* remains an identifier (Pfitzmann & Hansen, 2010, p. 2). Pragmatically and computationally, it can be more or less difficult to re-establish the link, depending for example on the background knowledge available to an attacker. The EU's General Data Protection Regulation (GDPR; European Union, 2016) on the one hand recognizes that pseudonymous data are not anonymous (and therefore require protection), and on the other hand, pragmatically, requires that it must at least be difficult to re-establish the link.[3]

The idea of difficulty must also be applied to the concept of anonymity because with unlimited background knowledge and computing power, it is rarely truly impossible to link messages to identifiers. The difficulty is operationalized by making definitions relative to the capabilities of an attacker. Thus, *computational anonymity* holds when an attacker subject to computational bounds cannot link an item (a message or a pseudonym) to an identifier. (Specific definitions of such bounds are the topic of attacker modelling.) In the remainder of this paper, we will treat "anonymity" and "computational anonymity" as synonymous.

The idea of *confidentiality* can be understood as coming from a complementary perspective: rather than focusing only on attackers that should *not* see some information, "preserving authorized restrictions on information access and disclosure […]"[4] emphasizes that there are intended and thus authorized recipients who should see it. Thus, if the information that Jane Doe said M is "kept confidential", this generally means that there is someone who is authorised to know that Jane said it, so this information exists. Therefore, it is stored somewhere (even if only in an ombudsperson's mind), maybe with linkage to Jane's real identity rendered more difficult by the use of pseudonyms, but not anonymous.

Current implementations of the (informal) notion of "anonymous whistleblowing support" often, in fact, involve information that is in fact pseudonymous. For example, a naïve system could offer a web entry form for the message that does not require a name – but log the IP address and time of message entry, which, together with the network connection data, in most cases uniquely identify a user. Systems based on Tor[5], on the other hand, send differently encrypted versions of the message through a network of computers in a way that makes it impossible to know, for any of these stations and for outside attackers, anything about the trajectory except the previous or next station, or what the message contains. Case credentials that support a conversation with the whistleblower that extends over time may be coupled to this connection data or otherwise be "just another pseudonym", or they may be generated by a more random process that – assuming standard computational bounds - allows the linkage only of the messages, but not of the message(s) and the sender. The functioning of each of these architectures, i.e. the level of anonymity that they can actually provide, is based on assumptions. These assumptions' credibility ranges from "unrealistic" to "high/state of the art" for Tor. In the next section, we will investigate these problems in more detail.

### 3.2 Limitations: communications anonymisation and epistemic non-anonymisability

Anonymity "can only be achieved if it is impossible for *any* person to identify the individual in question. Consequently, it is impossible to achieve anonymity in non-mediated interactions." (Marcum et al., 2020, p. 19, referring to Scott, 1998, emphasis added) Thus, because abused trust (in various actors) is what routinely breaks confidentiality, the quest for computational anonymity is always trying to not require trust in anybody.

Unfortunately, this is an ideal only; in real-life systems certain trust assumptions will always remain. For example, assumptions need to be made about how many nodes involved in transmitting a message are "honest" or "malevolent". Allowing more malevolent nodes involved, results in higher computational and/or communication overhead to preserve anonymity. For example, Tor functions even in very untoward settings with a high proportion of malevolent nodes. The sender selects three nodes that help to anonymize the communication and as long as not all three nodes collude malevolently, it functions (for details, see Dingledine, Mathewson, & Syverson, 2004). This comes at the cost that both communication and computation costs are quadrupled in comparison to an encrypted direct communication from sender to receiver.

---

[3] "pseudonymisation' means the processing of personal data in such a manner that the personal data can no longer be attributed to a specific data subject without the use of additional information, provided that such additional information is kept separately and is subject to technical and organisational measures to ensure that the personal data are not attributed to an identified or identifiable natural person" (Article 4 (5))

[4] https://csrc.nist.gov/glossary/term/confidentiality

[5] Torproject.org

Marx (1999) proposed a framework of threat types to anonymity, and Marcum, Young, & Kirner (2020) illustrated, with high-profile real-life examples, additional instances of these threats caused by today's electronic record-keeping and processing. We regroup and re-name them because the whistleblowing system's (WBS) technology plays different roles for these threats.

1. Some cues can lead to *direct re-identification*: in addition to unwitting disclosure of the whistleblower's legal name, these are easily linkable pseudonyms such as employee numbers or (on physical documents) fingerprints or DNA, or unwise choices by the whistleblower for case management data (username, password, secret credentials that are indicative of identity). These threats present database design and human-computer interaction challenges for WBS: pseudonyms should be system-generated to be secure. These challenges are technically solvable.
2. *Addresses* of various types (physical location, email address, telephone number, IP address, GPS coordinates) can make it simple to locate and reach the whistleblower (and then de-identify them). These threats present design challenges for WBS; for example, IP addresses should not be logged or even loggable. These challenges are technically solvable.
3. A wide range of surveillance technology may *track* the activities of everyone who comes into contact with the materials that are documenting evidence for whistleblowing, making it easy to re-identify them. This comprises various forms of logging, access control, and identifiers built into downloaded or printed documents (e.g. digital watermarks). Browser cookies may track if someone has documents open in one window and the WBS in another, and browser fingerprinting can identify the individual user. Cameras track physical movements. These threats can be difficult or impossible to overcome, especially since such security mechanisms may also or primarily be designed to track malicious insider behaviour, and since they are integral to the organisation's (or environment's) general operational architecture and thus not under the control of the WBS. The challenge here is that such logs might be part of otherwise needed security mechanisms. Hence, access to logs should be accordingly restricted to a minimum.
4. Report *metadata* (e.g. when a report was made, the voice of the reporter on a telephone hotline, the linguistic style and revealed lingo of a written report) may make it easy to infer demographics (e.g. gender), education, personnel category, etc., thereby reducing the size of the anonymity set. Computation may exacerbate this problem via machine-learning models that can predict demographics with high precision (e.g., Cesare et al., 2017). WBS design and procedures must be cognisant of this; ex-ante or ex-post schematisation of reports may help.
5. The perhaps largest problem is partially an inherently non-digital one: The very *content* of the message may be what leads to re-identification. This is a semantic and epistemic problem: Who can even know what is reported – for instance, if illegal behaviour was discussed in a specific meeting, it is known (including, often, to the reported-on) who participated. If only certain employees have access to or understanding of operational processes that form the core of the wrongdoing, then one of these must be at the root of the report. Again, the effect is an (often significant or down to 1) reduction of the size of the anonymity set. Again, We refer to this problem as *epistemic non-anonymisability*.[6] While ex post, instances of this problem cannot be solved by computational means (or not at all), there are ways of mitigating it, see Section 3.5.

---

[6] Kuhn et al. (2019) analyse a related problem formally and come to a similar conclusion. Their – theoretical – result is not directly applicable to the practical, real-life implementations relevant here; the implications of their work for our argument therefore need to be explored in future work.

Note that it may be desirable to design a whistleblowing channel in such a way that ex-post de-anonymisation is possible (e.g. by the whistleblower themselves, and only by them). Computationally, this is feasible, but beyond the scope of the present paper. We expand on this in (Schiffner, Berendt, & Schmitz, forthcoming).

### 3.3 Whistleblowing management as a service

The previous sections have addressed problems that are germane to all attempts to anonymise whistleblowing reports, whether these are internal to the organisation or external (usually via a media organisation or a government anti-corruption agency). This distinction between "internal" and "external" reflects a traditional form of service provision, in which all business/organisation functions are truly internal. In this section, we will concentrate on specific challenges that arise when organisations provide internal whistleblowing channels, but rely on external service providers to do so.

According to the WBD, legal entities in the private and public sector will have to establish channels and procedures for internal reporting and for follow-up (Article 8 (1)). These "shall enable the entity's workers to report information on breaches. They may enable other persons […] who are in contact with the entity in the context of their work-related activities to also report information on breaches." (Article 8 (2)). Channels must be secure to ensure confidentiality of the reporter and the reported-on, receipt must be acknowledged, an impartial person or department competent for following up on the report must be designated, communication to the reporter must be maintained (including, if applicable, follow-up questions and feedback to the reporter, and "diligent follow-up, where provided for in national law, as regards anonymous reporting"). Communication has to be conducted with diligence[7] and happen within specified or "reasonable" time frames. Also, "clear and easily accessible information regarding the procedures for reporting externally to competent authorities" must be provided (Article 9). These obligations translate into the requirement to have clearly set-up, implemented, documented, and non-trivial (because secure/confidential) procedures.

The channels and procedure can range from the informal and non-technical (employees talking to their superiors) via electronic communication means (e.g., email to an ombudsperson) to a full-fledged computational reporting platform. The latter approach corresponds to today's standards for software-supported company and public-sector processes, similar to bookkeeping or Human Resources. Such processes can be and often are outsourced. The WBD recognises that third parties can be authorised to receive reports of breaches on behalf of the original entity, provided that they offer appropriate guarantees of respect for independence, confidentiality, data protection and secrecy (Recital 54; Article 8 (5)). Such third parties can be external counsel, auditors, trade union representatives or employees' representatives, or – a common practice (de Sousa Costa & Ruivo, 2020) – external reporting platform providers.

These providers generally market their platforms as an "integrated solutions", thus relieving the client from the onerous obligations of system setup, basic communication such as acknowledging receipt, and/or documentation. "Solutions" range from software packages with support for the software (see for example the overview at https://www.capterra.com/whistleblowing-software/) to integrated hotline and reporting services (e.g., "Our world-class ethics hotline program offers professional specialists 24/7/365 with multiple language support, and our web intake and open door reporting options ensure you will capture more areas of misconduct to spot trends and take corrective action before minor issues become major.", https://www.navexglobal.com/en-us/products/hotline-reporting-and-intake). Vendors describe their systems as "completely anonymous, easy-to-use and fully customisable"[8] and as "an automated ethics and compliance solution"[9].

---

7 reasonable care or attention to a matter, which is good enough to avoid a claim of negligence, or is a fair attempt (https://legal-dictionary.thefreedictionary.com/diligent)

The existence of this market raises the question whether this constitutes "ethics as a service"[10] and if so, whether this amounts to an "ethics washing" or even an evasion of responsibility. We believe this to not be the case, since the outsourced services are basic bookkeeping activities, and the handling of the contents of reports and of the organisation's answer(s) to them remains the responsibility of the organisation. The organisation is also – at least under the GDPR – still fully legally responsible for ensuring the data protection of the personal data of the individuals involved, typically the reporter and the reported-on (de Sousa Costa & Ruivo, 2020). In other words, the truly ethical problems remain where they were: on the part of the employee, the dilemma between loyalty to the employer vs. to the law and/or higher values; on the part of the organisation, the holding to account of the wrongdoers, the rectification of the wrong, and the resistance to any temptations to retaliate – and of course the communication around it[11] [12]. In addition, it is possible that the non-functional requirements of security/confidentiality as well as anonymity guarantees are easier to meet for a third party specialising in these services rather than a small, multi-purpose internal IT department (cf. the long-standing debate whether or not to outsource cybersecurity, e.g. Pogrebna & Skilton, 2019).

With a view to anonymity, in-house solutions for anonymous channels might have a larger attack surface: First, it may be more likely that errors happen in the setting up and maintaining of anonymous channels when done in-house than when done externally, since specialised service providers often have superior general cybersecurity expertise. Second, certain signals that are given off inadvertently may lose their signalling power. Such signals might be recognized by human operators. For example, an external phone hotline operator is less likely to recognise the reporter's voice, and other metadata may also carry information and be cloaked by the external operator (such as the exact timing of a report, which may include or exclude certain employees from the anonymity set of reporters to those who can observe whether they are at work at that time),

There are, however, several concerns. First, like any software or architecture such systems and services could lead to subtle or less subtle behavioural changes both by management and employees. Second, given today's increasingly component- and service-based IT architectures, it is possible that information leakage occurs as an unintended consequence. For example, webservers create fingerprints in order to recognise returning users/devices. They do this to serve advertising, but also to recognise identity theft. As Adamsky et al. (2020) argue, it is today impossible to configure devices such that fingerprinting is avoided. The extent of both phenomena and possible countermeasures should be explored in future work.

A third concern is that most commercial providers of reporting platforms claim that they can handle anonymous reports and also establish back-channels (for giving feedback and for potential follow-up questions) – but they do not explain the underlying computational model. It may be that all is done is a non-logging of metadata such as IP addresses during communication and the provision of an identification token that the anonymous reporter can use for inquiring about their case. This does not provide sufficient guarantees, since simple hacking attacks (eavesdropping, man-in-the-middle) may lead to the re-identification of the reporter. Open-source systems such as GlobaLeaks or SecureDrop are built on Tor and provide much stronger guarantees. They do not require licensing fees, and by themselves do not provide the comfort of commercial integrated software solutions. Whether this adds up to cost savings or instead more expenditure, is a question every organisation must assess. It is noteworthy, however, that some large organisations, mostly but not limited to the public sector, have opted to use GlobaLeaks[13], and many (mostly media) organisations are using SecureDrop[14].

---

[8] https://www.capterra.com/p/211091/incy-io-Whistleblowing/

[9] https://www.navexglobal.com/en-us/resources/benchmarking-reports/2020-regional-whistleblowing-hotline-benchmark-report?RCAssetNumber=8053

[10] We thank Geoffrey Rockwell for devising this term.

[11] cf. for example "Whistleblower Software enables confidential or anonymous two-way communication between whistleblowers, and the people assigned to deal with reports inside an organization." https://www.capterra.com/whistleblowing-software/

[12] e.g., an "Ethics & Controls Compliance" Director: "NAVEX Global redirects issue reports to the proper department and allows us to assign and follow up on cases electronically. It has saved a tremendous amount of time, which allows me to do what I should be doing: conducting investigations and answering inquiries." cited at https://www.navexglobal.com/en-us/products/hotline-reporting-and-intake

### 3.4 How is the "anonymous" whistleblower protected legally?

So if anonymity cannot be guaranteed technically and retaliation cannot be prevented in all cases, how does the law protect supposedly anonymous whistleblowers? We will discuss this in an exemplary fashion with respect to European law.

In a report by the Italian anti-corruption agency ANAC on received and processed whistleblowing, we find the following passage: "… the [percentage] of anonymous reports received by the interviewed [administration employees] is substantial; these reports are generally processed, whilst bearing in mind that there is no whistleblower to protect."[15] This is a peculiar finding. ANAC has been involved in a lengthy development and discussion process resulting in the use of a variant of GlobaLeaks (Steph, 2018), and therefore would be expected to be quite knowledgeable about anonymisation. The passage may just be unfortunate wording, but it may also reveal a fundamental uncertainty on the part of the persons actually dealing with "anonymous" reports: Does an "anonymous" reporter exist in the sense of deserving protection?

Consider first the non-anonymous case. Whistleblowing reports in general start their "lives" as personal data and as such require the technical and organisational measures appropriate to protect them as well as their data subjects (the reporter as well as the reported-on). Any subsequent confidentiality applied, e.g. when the persons tasked with handling the reports hide the reporter's identity from management, does not invalidate this feature. The GDPR mandates the use of state-of-the-art methods and fairness (Article 5 (1)(a)), but procedures for determining the state of the art do not yet exist (see Schiffner et al., 2018). In addition, anonymous data do not count as personal data and are not protected under the GDPR (Recital 26).[16] Therefore, it may be argued that data that are anonymous from the outset, rather than anonymized ex post, never fall under the protection of the GDPR.

The WBD is clear: "Persons who reported or publicly disclosed information on breaches anonymously, but who are subsequently identified and suffer retaliation, shall nonetheless qualify for the protection provided for under Chapter VI, provided that they meet the conditions laid down in paragraph 1." (Article 6(3)). We interpret this to mean that the WBD considers data from anonymous reports as, in principle, "anonymous only in a preliminary way". In addition, all modern whistleblowing-support systems provide anonymous reporters with case credentials or otherwise enable them to later prove that they made the report. In a computational sense, this amounts to providing them with a pseudonym (generally one valid even across different transactions, to allow back-and-forth follow-up).

Insisting on the status of these data as anonymous would present conflicting incentives for those controlling and processing the data: should they do their best to cloak whatever potentially identity-revealing information remains in a report and its metadata in order to avoid de-anonymisation (and forego both GDPR duties on themselves and GDPR protections for the reporter), or should they do their best to protect these data as if they were already personal data[17]? We consider it in line with the spirit of the GDPR and the WBD to prefer the second option.

---

[13] E.g., https://www.globaleaks.org/usecases/anti-corruption/, https://www.globaleaks.org/usecases/investigative-journalism/

[14] https://securedrop.org/

[15] "Il monitoraggio riferito alle amministrazioni mostra anche il numero di segnalazioni anonime che i soggetti intervistati hanno ricevuto che, in termini percentuali, risulta cospicuo; queste segnalazioni vengono in genere trattate, pur nella consapevolezza che non c'è alcun *whistleblower* da tutelare." (ANAC, 2019, p. 6)

[16] It should be kept in mind here that pseudonymised data *do* fall under the protection of the GDPR (Recital 26, Article 4 (5)).

Thus, we conclude that not only semantically (see Section 3.2) but also legally, "anonymous" whistleblowing reports are usually at least pseudonymous data with respect to the reporter, and as such are protected under the GDPR. And since the ultimate purpose of the GDPR is not to protect data, but persons, the "anonymous" whistleblower exists and has a right to protection under both the WBD and the GDPR.

However, due to the still-unclear definitions, it is not clear whether the use of methods that, from a computational standpoint, clearly provide fewer guarantees (see Section 3.2), is a violation of the state-of-the-art requirement in the GDPR legal sense. In any case, it appears unfair to not inform the "users", in particular the potential whistleblowers, of the possible ways in which their data may be de-anonymised.[18]

### 3.5 Governance beyond technology and law

The governance strategies outlined in the previous sections are high-level, and neither of them is guaranteed to work. How they play out in real-life contexts will depend on many further choices in organisations, and personal morality, professional codes of conduct, as well as company-specific or sectoral ethics codes can play a role. They can be a source of normativities beyond the law and help to instantiate knowledge and values into managerial choices. Epistemic non-anonymisability, mentioned in Section 3.2, may appear to be a categorical, given, and unsolvable problem that threatens to invalidate technical approaches wholesale. However, this is not the case. Professionals should understand that epistemic non-anonymisability may also be the result of choices including managerial and IT settings. For example, anonymity sets will have size one (or another small number) if meetings may contain a very small number of persons, if access to electronic resources is restricted and/or logged, if printouts are watermarked, or if employees are tracked.[19] In addition, managerial and legal settings can affect epistemic non-anonymisability: WBS procedures that encourage, ex-ante, an 'as coarse as possible' level of detail, or that involve ex-post transformation of reports may be useful, but the information must be specific enough to constitute a degree of suspicion sufficient to initiate legal pursuit.

Codes specify *values* and *norms*. For example, the ACM/IEEE Code of Conduct for Software Engineers (Gotterbarn, Miller, & Rogerson, 1997) requires "fairness and being supportive" to colleagues in addition to "acting in the best interest of" employers and clients (and the public interest), and a small number of the currently many "AI ethics" codes require developers and organisations to protect ethical whistleblowers (Whittaker et al. 2018, IEEE 20161; see overview in Hagendorff, 2020). These values and norms need to be complemented by the requirement for software engineers to have *knowledge* about the effects of logging (etc.) on anonymity sets. On the one hand, this strengthens the case for choices (such as the restriction or coarsening of logging) that are normally only considered to be protecting the privacy of employees. On the other hand, it is questionable what real impact codes of conduct will have, given that – certainly in the area of computing and AI – lack enforcement mechanisms and can bring software engineers into conflict with managerial power (Hagendorff, 2020, Haas, Gießler, & Thiel, 2020). Their impact may grow once ethics codes become law.[20]

---

[17] Veale, Binns, and Ausloos (2018) have analysed this dilemma in a more general form.

[18] While unfair, this may not be illegal, since Article 13 GDPR does not oblige controllers to inform data subjects on security measures.

[19] The NSA whistleblower Reality Winner was de-anonymised based on a combination of logging, watermarking, and media carelessness (Wemble, 2017).

[20] For example, the EU "Guidelines for Trustworthy AI" were a major basis for the current proposal for a (binding) regulation of AI. Article 14 (4) of this "AI Act Proposal" requires mechanisms for human oversight of AI decision-making to ensure that users are aware of automation bias, i.e. the empirically well-documented tendency of human "overseers" to blindly follow the machine's guidance. This is an attempt to ensure knowledge of specific limitations of technology-based systems among users. In analogy with this provision, laws could require WBS users to be made aware of limitations with regard to their anonymity, and the designers of all systems in organisations to

Still, just like legal provisions, codes of conduct and similar norms are not guaranteed to deter retaliation. Could retaliation be made *unattractive* (rather than forbidden or impossible)? Retaliation is attractive, among other reasons[21], because it is shameful to have made an error (and be called out for it by a whistleblower). Can strategies be found to address this root cause? Acknowledging errors is necessary for the development of an ethics of responsibility in organisations and society (Weingardt, 2004), but it can also have very practical advantages.

Quality assurance systems in workplaces (e.g. the Japanese Kaizen, cf. Colenso, 2000), the health sector, and aviation have highlighted another important factor: the valuation of the reported-on actions and the report itself. Is what happened considered a deficit or error and the reporting perceived to be an opportunity for the system to learn and avoid such errors in the future? Or is it considered a misdeed, perceived as a (maybe permanent) loss of opportunity for the reported-on or the system as a whole to function, prosper, etc.? The success of Critical Incident Reporting Systems indicates that an appropriate "*error culture*"[22] can contribute to reporting and organisational learning (Schüttelkopf, 2008). It is likely that organisational cultures that strive only for (even if seeming) flawlessness and perfection, via the pressure to conform to this ideal, deter reporting.[23] It has to be acknowledged, however, that some wrongs cannot be cast as an error, that calling criminal intent by its name is sometimes required.

The possibility to remain anonymous is also regarded as key for Critical Incident Reporting Systems (Flanagan, 1954), in other words, the lifting of the fear of sanctions may impact the likelihood of reporting independently of whether wrongs are considered errors or misdeeds. In the limit, a functioning error culture could reduce or obviate not only the 'need' to retaliate, but also the 'need' to blow the whistle. In sum, anonymity (mainly enabled by *technical* means), *legal* and other *norms* (accompanied by practices and sanctions), and *organisational* culture are possible and often necessary complements to one another. We call this the "triangle of whistleblowing protection and incentivisation".

## 4 Summary, conclusions, and outlook

In this paper, we have investigated how computational technology can help with, and affects, ethical issues linked to whistleblowing. In particular, we have analysed the possibilities and limitations of anonymization.

*First*, we have argued that only software can effectively provide anonymity, but to do so, it must implement state-of-the-art anonymous-communications technology and transparency requirements. Open-source systems that do this are already successfully deployed by companies and governmental agencies. *Second*, there are theoretical, principled limitations to anonymity: communications technology cannot guarantee anonymity for all reports because the content may reveal identity; we call this "epistemic non-anonymisability".

*As a consequence*, it is dangerous (and not in the interests of potential whistleblowers) to rely on the possibility of "anonymous" reporting and/or improvements to legal or organisational anonymity practices alone. Instead, we proposed a "triangle of whistleblowing protection and incentivisation": (a) anonymity in a formal and

---

be made aware of the implications of logging etc. for whistleblowing. It should be noted, however, that such awareness and the transparency enabling it are, by themselves, not sufficient to guarantee success, e.g. if they conflict with power (Berendt, 2022b).

[21] Other reasons such as being vengeful or knowing that the reported-on behaviour was indeed criminal and does not allow for any exculpation, are outside the scope of this subsection.

[22] In German: *Fehlerkultur. The term is unwieldy already in German, and intentionally so: the concepts of "culture" and "errors" seem to clash and we have to make a conscious effort to bring them together. For this reason, we use a literal translation.*

[23] Westin (1981, p. 148) provides indirect support for this conjecture: participatory work environments "provide an atmosphere for raising problems that is more hospitable to resolving whistle-blowing charges than the hierarchical management tradition."

technical sense, (b) whistleblower protection (through specific whistleblower-protection legislation per se *and* in interaction with other laws), and (c) other norms and practices including codes of conduct as well as organisational error culture (comprising the organisation's perspective on misconduct, actions are taken in response to reports, and these actions harmful or beneficial effects on the different stakeholders of reports). Design and implementation of these mechanisms requires the collaboration of (at least) experts in law, technology and social sciences. This collaboration will serve companies' bottom line and also the Common Good.

Many questions remain for future work. We want to highlight (i) the need for more interdisciplinary investigations into how to reduce instances of epistemic non-anonymisability and (ii) four questions that follow from the recent developments of whistleblowing support as a service. First, what behavioural changes, of management, employees, and other stakeholders does such outsourcing create? Second, how secure and "private" are these services, and especially how secure/private relative to whom? Role distribution and threats have to be analysed, on a theoretical as well as on a practical level. Moreover, monetary and intangible assets need to be identified and their threat exposure needs to be estimated in order to allow a quantification of Risks. Third, with what empirical methods can all these questions be investigated; to what extent do formal analyses and laboratory experiments generalize to real-life settings? Fourth, what ethical challenges arise from answers to these questions?

## Acknowledgements

We thank Rita de Sousa Costa and Inês de Castro Ruivo for a great talk on whistleblowing in the Ethics Dialogues 2021 that motivated us to expand on our initial ideas for this paper, and for further invaluable discussions and feedback. We are also grateful to all the other participants of the Symposium on Ethics in the Age of Smart Systems 2021 for great discussions. Bettina Berendt received funding from the German FederalMinistry of Education and Research (BMBF) – Nr. 16DII113f. Stefan Schiffner was supported by the Luxembourg National Research Fund as part of EnCaViBS, grant number C18/IS/12639666/EnCaViBS/Cole.